\documentclass{article}
\usepackage{spconf,amsmath}
\usepackage{cite}
\usepackage{hyperref}

\usepackage{graphicx}
\usepackage{newfloat}
\usepackage{xcolor}
\usepackage{siunitx}
\usepackage{collcell}
\usepackage{graphicx} 

\usepackage{tikz}
\usepackage{subcaption}
\usepackage{mathtools}
\usepackage{url}
\usepackage[font=small]{caption}
\usepackage[utf8]{inputenc}         
\usepackage{wrapfig}

\usepackage[full]{complexity}

\usepackage{polski}
\usepackage{xcolor}

\usepackage[T1]{fontenc}  
\usepackage{amsmath,amsfonts,amssymb,amsthm}

\begin{document}
\title{Cloud Detection in Multispectral Satellite Images Using Support Vector Machines With Quantum Kernels}

\name{\begin{tabular}{c}Artur Miroszewski$^{1}$, Jakub Mielczarek$^1$, Filip Szczepanek$^1$, Grzegorz Czelusta$^1$,\\{Bartosz Grabowski}$^{2}$, \textit{Bertrand Le Saux}$^3$, and \textit{Jakub Nalepa}$^{2,4}$\end{tabular}}

\address{$^1$Jagiellonian University, prof. S. \L{}ojasiewicza 11, 30-348 Cracow, Poland\\$^2$KP Labs, Konarskiego 18C, 44-100 Gliwice, Poland\\$^3$European Space Agency, Largo Galileo Galilei 1, 00044 Frascati, Italy\\$^4$Silesian University of Technology, Akademicka 16, 44-100 Gliwice, Poland
\\ \texttt{artur.miroszewski@uj.edu.pl, jnalepa@ieee.org}}

\maketitle              
\begin{abstract} 
Support vector machines (SVMs) are a well-established classifier effectively deployed in an array of pattern recognition and classification tasks. 
In this work, we consider extending classic SVMs with quantum kernels and applying them to satellite data analysis. 
The design and implementation of SVMs with quantum kernels (hybrid SVMs) is presented. It consists of the Quantum Kernel Estimation (QKE) procedure combined with a classic SVM training routine.
The pixel data are mapped to the Hilbert space using ZZ-feature maps acting on the parameterized ansatz state. The parameters are optimized to maximize the kernel target alignment.
We approach the problem of cloud detection in satellite image data, which is one of the pivotal steps in both on-the-ground and on-board satellite image analysis processing chains. The experiments performed over the benchmark Landsat-8 multispectral dataset revealed that the simulated hybrid SVM successfully classifies satellite images with accuracy on par with classic SVMs.

\end{abstract}
%
%
\section{Introduction}

Satellite imaging plays an increasingly important role in various aspects of human activity. The spectrum of applications ranges from cartographic purposes~\cite{Copernicus-Land,castillo2021semi} through meteorology~\cite{weather-forecast-satellites}, ecology, and agronomy~\cite{Nalepa2022hyperview} to security and urban monitoring~\cite{audebert2017deep}. Consequently, dozens of terabytes of raw imaging data are generated daily from satellite constellations, such as the one built within the European Copernicus Programme. The large volume of multi- or hyperspectral images, which capture the detailed characteristics of the scanned materials, makes them difficult to transfer, store, and ultimately analyze. Therefore, their reduction through the extraction of useful information is a critical issue in real-world applications. An important step in the data analysis chain of optical satellite data is the identification of clouds. The interest is two-fold: on the one hand, such cloudy regions can be removed from further processing, as the objects of interest are likely to be obscured. On the other hand, efficient detection of cloud cover on the Earth surface is important in meteorological and climate research~\cite{9554170}. 

Since the reduction is performed on a huge amount of raw data, the efficiency of this process is a key factor in practice. Therefore, it is reasonable to search for new methods for analyzing such huge datasets, improving image data classification into cloudy and clear areas. 
In this paper, we investigate the classification performance of classic SVMs exploiting the radial basis function kernels, and those which benefit from the quantum kernels (introduced in Section~\ref{sec:materials_methods}). There are theoretical arguments \cite{havlivcek2019supervised,goldberg2017complexity,demarie2018classical} that the proposed quantum kernels are \#\P-hard to evaluate on a classical computer. Therefore, if they provide an advantage in classification accuracy, this would advocate a strong use case for quantum computers.
Additionally, to get a deeper understanding of quantum machine learning mechanisms and show its usefulness in practice, it is pivotal to focus on widely adopted image data corresponding to real use cases. 
Thus, we tackle the cloud detection task in satellite image data, which is one of the most important processing steps for such imagery. 
Our experimental study performed over the benchmark multispectral image data acquired by the Landsat-8 satellite revealed that SVMs with quantum kernels offer a classification accuracy at least comparable to classic RBF kernel SVMs (Section~\ref{sec:experiments}). 

\section{Materials And Methods}\label{sec:materials_methods}
\subsection{Data}
We utilize satellite image data contained in the 38-Cloud dataset~\cite{38-cloud-2}.
The data consist of Landsat 8 scene images cropped into $384 \times 384$ pixel patches.
Each pixel has five values associated with it: intensity values in four spectral bands (blue: 450--515 nm, green: 520--600 nm, red: 630--680 nm, NIR: 845--885 nm) and a label corresponding to the fact whether a pixel contains a cloud or not. Therefore, the dimension of the data is $m=4$.
SVMs suffer from their high time and memory training complexity, which depend on the size of the training set. Since only a subset of all training vectors is annotated as support vectors during SVM training, we can effectively exploit only a subset of the most important examples~\cite{Nalepa2019}.
To find the best training data, two metrics for patches were introduced:
cloudiness $\mathcal{C}$ (the ratio of cloud pixels to all pixels in the
patch), fill $\mathcal{F}$ (the ratio of physical pixels in the patch, as patches contain scene margins). Balanced training sets from patches with properties $\mathcal{F} = 100\%$, $ 40\% \leq \mathcal{C} \leq 60\%$ are sampled by randomly selecting a fixed number of pixels.
\subsection{Methods}
\begin{figure}[ht]
\centering
\includegraphics[width=0.5\textwidth]{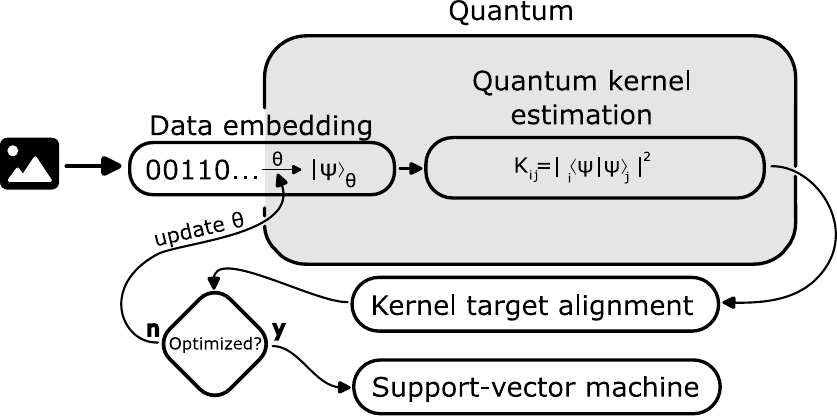}
\caption{Overview of the hybrid SVM design. The gray box indicates the part of the algorithm which was calculated on the quantum computer simulator.} \label{fig:design}
\end{figure}

In Fig. \ref{fig:design}, we present a high-level flowchart of the proposed hybrid SVM procedure. First, we encode the data with the parameterized feature map consisting of ZZ-feature map acting on ansatz (see Fig. \ref{fig:ansatz}). Then we perform the Quantum Kernel Estimation (QKE) and obtain a quantum kernel, which target alignment is maximized with a classic optimization method. When optimization is finished, the final quantum kernel is passed to a classic SVM.

\subsubsection{Selected Feature Map}
When considering the employment of quantum computation methods, a principal question that
quickly arises pertains to the way in which classic input data will be loaded into the quantum circuit.
In general, our aim will be to construct a unitary operator for each input datum $x$, such that applying it to the
initial quantum zero state will leave us with a specified representation of $x$, $|\phi(x)\rangle$.
This process is called \textit{quantum embedding}, while any such map $x \mapsto |\phi(x)\rangle$ is known as a \textit{quantum feature map}.

Consider the unitary transformation 
\begin{equation}
    U_{\phi(x)} = \exp \left( i \sum_{S\subseteq [n]} \phi_S(x) \prod_{i \in S} P_i \right),
\end{equation}
being a general quantum circuit Pauli expansion of an $n$-qubit unitary transformation.
The index $S$ describes the connectivities between different qubits: $S \in \{ {n \choose k}\ \text{combinations}, k\in \{1, \dots, n\} \} $, $P_i$ are the basic Pauli gates that act on the $i^{th}$ quantum register and the data mapping function is $\phi_{\{i\}}(x) = \pi x_i$, $\phi_{\{i,j\}}(x) = \pi (1-x_i)(1-x_j)$. The number of qubits used can be identified with the dimension of the data $n=m$.
Following \cite{havlivcek2019supervised}, we restrict the above unitary to $k = 2$ connectivities with single-qubit gates $P_{\{a\}} = Z_a$, two-qubit gates $P_{\{b,c\}} = Z_b Z_c$, $a,b,c \in \{0,\dots, n-1\}$.
This transformation is called a ZZ feature map with one repetition. It is already $\#\P$-hard to calculate classically \cite{goldberg2017complexity}, but shows no computational advantage over the classical kernel estimation, performed by random sampling \cite{demarie2018classical}.
To increase the complexity of the classical simulation of the ZZ feature map, additional bases-changing layers are included by repeating the $\tilde{U}_{\phi(x)}$ transformation
\begin{equation}
    \mathcal{U}^d_{\phi(x)} = \left( \tilde{U}_{\phi(x)} \right)^d = \left( U_{\phi(x)} H^{\otimes n} \right)^d, d \in \mathbb{N}.
\end{equation}
The transformation $\mathcal{U}^d_{\phi(x)}$ is called the ZZ feature map with $d$ repetitions.

\subsubsection{Circuit Parameterization}
\begin{figure*}[t]
    \centering
    \includegraphics[width = 0.9\textwidth]{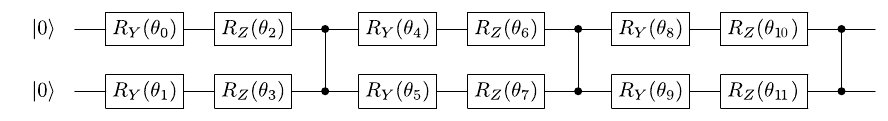}
    \caption{The ansatz $V_{\theta} | 0  \rangle^{\otimes 2}$ circuit, which prepares the initial state for the kernel estimation for classical data of dimension $m=2$.}
    \label{fig:ansatz}
\end{figure*}
Having defined the ZZ feature map, we turn our attention to the possibility of introducing additional parameters into the circuit to maximize the kernel target alignment of the data in the feature space.
We follow the approach of modifying the initial state of the circuit \cite{glick2021covariant} on which the feature map acts.
The initial state will be parameterized with continuous variables $\theta$ (Fig. \ref{fig:ansatz}), with respect to which we will perform kernel target alignment optimization. 

\subsubsection{Kernel Target Alignment}
Considering a collection of quantum states obtained by means of applying a quantum embedding $|\phi(x)\rangle = \mathcal{U}_{\phi(x_i)}\ V_{\theta} | 0  \rangle^{\otimes 2}$ to different classic input data, it is straightforward to reason about them in terms of kernel methods.
The kernel $\mathcal{K}$ with respect to any two embedded data $x_i$, $x_j$ can be defined as the fidelity between the resulting quantum states,
\begin{equation}
\label{eq:quantum-kernel}
    \mathcal{K}_{ij} = |\langle \phi(x_i) | \phi(x_j) \rangle|^2 = |\langle 0 |^{\otimes n}\  V_{\theta}^{\dagger} \ \mathcal{U}^{d\ \dagger}_{\phi(x_j)} \mathcal{U}^d_{\phi(x_i)}\ V_{\theta} | 0  \rangle^{\otimes n}|^2.
\end{equation}
This kernel $\mathcal{K}$ is known as the \textit{quantum kernel}.
Consider a kernel function given by
\begin{equation*}
    \bar{\mathcal{K}}_{ij}= \begin{cases}
    +1 &\text{if $x_i$ and $x_j$ are in the same class}\\
    -1 &\text{if $x_i$ and $x_j$ are in different classes}.
    \end{cases}
\end{equation*}
It shows a clear distinction between classes of data points and is called an \textit{ideal} kernel.
In general, in almost every situation, one will not be able to find the exact feature map, which gives rise to the ideal kernel. 
Therefore, parametrized families of feature maps are used to optimize the resulting kernel matrix in such a way that it resembles the ideal kernel as closely as possible.
The function that indicates the similarity between a specific and ideal kernel matrices is called kernel target alignment \cite{cristianini2001kernel}
\begin{equation}\label{eq:target_alignment}
\mathcal{T}(\mathcal{K}) = \frac{\langle \mathcal{K}, \bar{\mathcal{K}} \rangle_F}{\sqrt{\langle \mathcal{K}, \mathcal{K} \rangle_F \langle \bar{\mathcal{K}}, \bar{\mathcal{K}} \rangle_F}},
\end{equation}
where $\langle A, B\rangle_F = Tr\{ A^T B \}$ is a Frobenius inner product.

\section{Experimental Results}\label{sec:experiments}
The objective of our study is to compare hybrid SVMs with their classic counterparts.
The results, presented in Table \ref{tab:results}, are obtained by using the \verb|Qiskit| Aer simulator, whereas the optimization algorithm is the standard simultaneous perturbation stochastic approximation.
The SVM score is obtained from the \verb|sklearn| support vector classification, with the radial basis function (RBF) kernel with default $\gamma = \frac{1}{m\sigma^2}$ (where $m=2$ is the number of features and $\sigma^2$ is the variance of the data) and $C = 1$, the latter being the regularization parameter. For each simulation run, we randomly sample 800 pixels for the training set, and 200 pixels for the test set (the training and test sets are non-overlapping). 
To keep the size of the quantum circuit minimal, we decrease the number of data features by running a principal component analysis before feeding it into the algorithm. The number of four spectral bands is reduced to two features. 
The ZZ-feature map is chosen to consist of $d=2$ repetitions. The results of the Wilcoxon matched-pairs signed rank test show that there is no statistically significant difference between hybrid SVMs with a quantum kernel and classic SVMs with an RBF kernel (at $p<0.05$). Therefore, it shows that the classification methods with quantum kernels based on ZZ-feature map are, at least, competitive with classical SVM models. With the further development of quantum kernels, we expect hybrid methods to be advantageous over classical classification methods. 
\begin{table*}[ht]
\caption{The results of the circuit simulations on 20 different training-test 38-Cloud splits: hSVM and SVM indicate classification accuracy for hybrid SVM and classic SVM methods. $\mathcal{T}_i$ and $\mathcal{T}_f$ show kernel target alignment before and after optimization.}\label{tab:results}
\centering
\begin{tabular}{lcccclcccc}
\hline
             &                     &                   & \textbf{$\mathcal{T}_i$} & \textbf{$\mathcal{T}_f$}                  & \textbf{hSVM} & \textbf{SVM}       &                   &               &               \\ \hline
\multicolumn{3}{l}{\textbf{Average}}                   & 0.049               & 0.081                              & \multicolumn{1}{c}{0.778}         & 0.788               &                   &               &               \\
\multicolumn{3}{l}{\textbf{Standard deviation}}                       & 0.018               & 0.024                              & \multicolumn{1}{c}{0.038}          & 0.029               &                   &               &               \\ \hline
\end{tabular}
\end{table*}

\section{Conclusions}\label{sec:conclusions}
We introduced the design and implementation of an SVM with quantum kernels. 
The proposed algorithm was experimentally verified on the cloud detection benchmark dataset. 
The main takeaway from the work is that---at this stage---SVMs with the quantum kernel have a classification accuracy on par with classic SVMs with RBF kernel.
The experiment was performed with a quantum computer simulator.
In \cite{miroszewski2023detecting} the use of underparametrized quantum circuits for similar task was performed on the whole 38-Clouds data set. Current results are consistent with this work.
To estimate the effect of noise and better understand the computational time, more work should focus on running the algorithm on quantum computers. 
We anticipate that further investigation of the quantum feature maps---including using full dimension of the dataset ($m=4$), new data mapping functions or different generators---will result in an additional improvement of the classification performance. 
This would indicate a strong use case for quantum computers in SVM models.


%
%
%
\bibliographystyle{IEEEbib}
\bibliography{bibliography}
%





\subsubsection*{Acknowledgements} 

This work was funded by the European Space Agency,
and supported by the ESA $\Phi$-lab 
(\url{https://philab.phi.esa.int/}) AI-enhanced 
Quantum Computing for Earth Observation (QC4EO) 
initiative, under ESA contract No. 4000137725/22/NL/GLC/my. 
AM, JM, GC, and FS were supported by the Priority Research 
Areas Anthropocene and Digiworld under the program 
Excellence Initiative – Research University at the 
Jagiellonian University in Krak\'ow. JN was supported 
by the Silesian University of Technology grant for 
maintaining and developing research potential.

\end{document}